\documentclass[letterpaper]{article} 
\usepackage{aaai2026}  
\usepackage{times}  
\usepackage{helvet}  
\usepackage{courier}  
\usepackage[hyphens]{url}  
\usepackage{graphicx} 
\usepackage{natbib}  
\usepackage{caption} 
\usepackage{algorithm}
\usepackage{algorithmic}

\usepackage{amsmath}
\usepackage{amsfonts}
\usepackage{booktabs}
\usepackage{multirow}
\usepackage{colortbl}
\usepackage{makecell}
\usepackage{tabularx}

\usepackage{newfloat}
\usepackage{listings}
\DeclareCaptionStyle{ruled}{labelfont=normalfont,labelsep=colon,strut=off} 
\floatstyle{ruled}
\newfloat{listing}{tb}{lst}{}
\floatname{listing}{Listing}
\title{Temporally Decoupled Diffusion Planning for Autonomous Driving}
\author{
Xiang Li\textsuperscript{\rm 1},
Bikun Wang\textsuperscript{\rm 1}\thanks{Corresponding author.},
John Zhang\textsuperscript{\rm 1},
Jianjun Wang\textsuperscript{\rm 1}
}
\affiliations{
    \textsuperscript{\rm 1}Bosch Cross-Domain Computing Solutions\\

    wodlxwhy@gmail.com,Bikun.WANG@cn.bosch.com,external.John.ZHANG@cn.bosch.com,Jianjun.WANG2@cn.bosch.com
}

\usepackage{bibentry}

\begin{document}

\maketitle

\begin{abstract}
Motion planning in dynamic urban environments requires balancing immediate safety with long-term goals. While diffusion models effectively capture multi-modal decision-making, existing approaches treat trajectories as monolithic entities, overlooking heterogeneous temporal dependencies where near-term plans are constrained by instantaneous dynamics and far-term plans by navigational goals. To address this, we propose Temporally Decoupled Diffusion Model (TDDM), which reformulates trajectory generation via a noise-as-mask paradigm. By partitioning trajectories into segments with independent noise levels, we implicitly treat high noise as information voids and weak noise as contextual cues. This compels the model to reconstruct corrupted near-term states by leveraging internal correlations with better-preserved temporal contexts. Architecturally, we introduce a Temporally Decoupled Adaptive Layer Normalization (TD-AdaLN) to inject segment-specific timesteps. During inference, our Asymmetric Temporal Classifier-Free Guidance utilizes weakly noised far-term priors to guide immediate path generation. Evaluations on the nuPlan benchmark show TDDM approaches or exceeds state-of-the-art baselines, particularly excelling in the challenging Test14-hard subset.
\end{abstract}

%
\begin{links}
    \link{Code}{https://github.com/wodlx/TDDM}
\end{links}

\section{Introduction}

Motion planning is one of the core technical challenges in autonomous driving~\cite{chen2024e2e, aradi22reinforce}, with the objective of generating safe, comfortable, and human-like trajectories in dynamically changing and uncertain environments~\cite{chen2024end}. Traditional planning methods, such as search-based~\cite{stilman2008planning} or optimization-based approaches~\cite{fan2018baidu}, rely on precise environmental models and complex heuristic rules~\cite{treiber00phy, dauner23apdm}. Although they can be highly efficient and perform well in scenarios covered by these rules, they often struggle to cope with the highly dynamic and interactive nature of complex urban scenarios~\cite{mitduc2009}.

\begin{figure}[t]
\centering
\includegraphics[width=0.98\columnwidth]{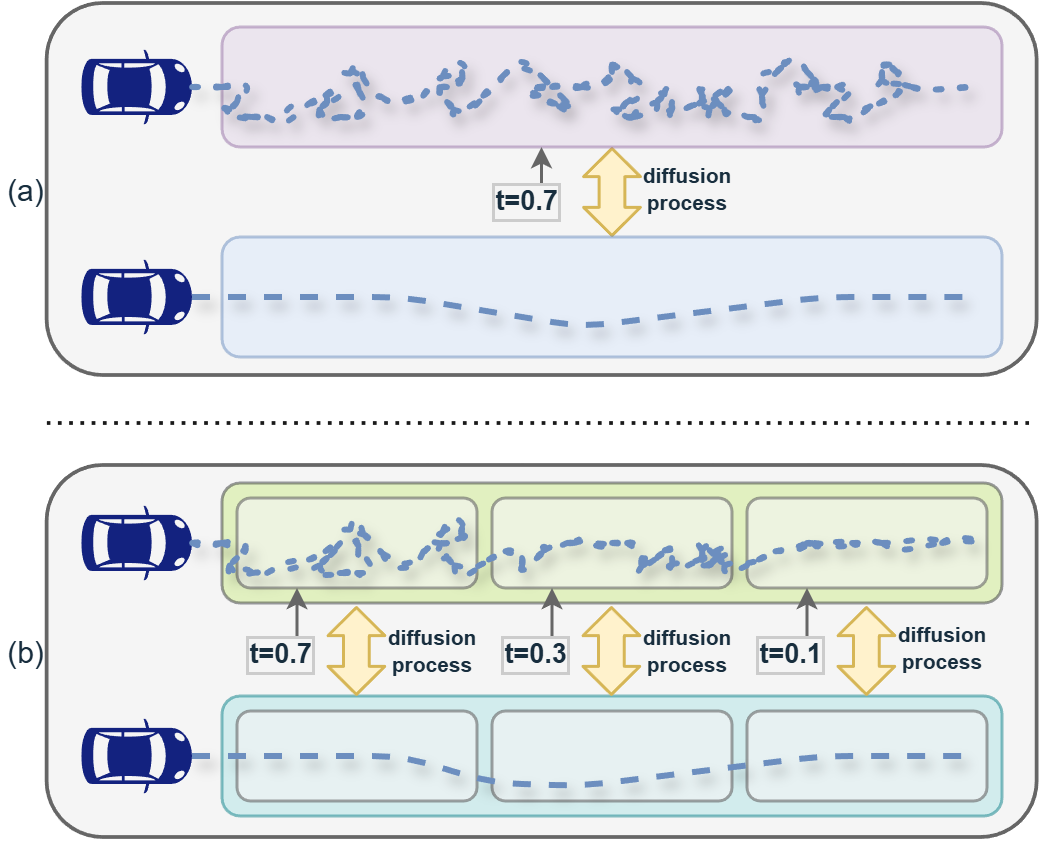} 
\caption{\textbf{The comparison of our temporal decoupled diffusion model and full sequence diffusion model. }(a) full sequence diffusion model. (b) temporal decoupled diffusion model (Ours). It can be seen that the biggest difference between our model and the full sequence model is the support for independent diffusion processes on temporal segments, where $t$ denotes the diffusion timestep.}
\label{fig1}
\end{figure}

In recent years, learning-based planning has emerged as the mainstream paradigm. By learning driving policies from large-scale data, these methods can better generalize to complex and variable real-world scenarios, reducing the reliance on hand-crafted rules. Among learning-based approaches, imitation learning (IL)~\cite{pomerleau1988alvinn, bansal2018chauffeurnet}, especially end-to-end (E2E) methods~\cite{hu2023planning,cheng2024pluto}, has achieved significant success. By directly learning the mapping from perception to control, it greatly simplifies the complexity of the traditional planning framework. However, classic imitation learning methods face two core challenges. The first is \textit{distribution shift}~\cite{cheng2023plantf}, where the state distribution encountered during testing differs from the training data, leading to compounding errors. The second is the difficulty in capturing the inherent \textit{multi-modal decision-making}~\cite{cui2019mtp, chen2024vadv2} characteristic of real-world driving scenarios. Faced with the same situation, a human driver might make several equally valid and safe decisions, whereas a single regression target often yields overly conservative or averaged-out policies.

To resolve this, recent literature shifts from deterministic regression to generative sequence modeling. To align with the broader automated planning community, it is crucial to distinguish these paradigms: classical planning searches for an optimal action sequence based on explicit logical constraints, whereas trajectory generation reframes the task as conditional probability modeling. Instead of explicit logical search, generative models learn to sample diverse, contextually plausible future trajectories from a learned data distribution.

Driven by this generative formulation, diffusion models~\cite{ho2020denoising, song2021maximum} have been introduced into the field of trajectory planning, owing to their powerful ability to model complex data distributions and their flexibility in being guided by different conditions to achieve various goals. By redefining the planning task as a conditional trajectory generation problem, diffusion models can start from random noise and, through a denoising process guided by scene context and kinematic constraints, generate a diverse set of plausible trajectories that are consistent with the scene. The success of applying diffusion models to trajectory generation has been demonstrated in several works~\cite{jiang2023motiondiffuser,liao2025diffusiondrive,zheng2025diffusionbased}. However, existing diffusion-based planning methods often treat the entire trajectory as a monolithic, indivisible entity~\cite{zheng2025diffusionbased}. This ignores the varying dependencies of different temporal segments of the trajectory in the scene context. For instance, the initial segment of a trajectory (near-term plan) is more dependent on the vehicle's precise current state and the instantaneous dynamics of the surrounding environment, whereas the final segment (far-term plan) is more focused on achieving long-term navigational goals.

Inspired by recent advances in sequential modeling from the video generation domain, we draw an analogy between trajectory generation and the generation of video frame sequences. Cutting-edge methods in video generation, such as Diffusion Forcing~\cite{chen2025diffusion}, achieve higher quality and longer-term video extrapolation by applying independent noise to different frames and modeling their temporal relationships. Similarly, works like CausVid~\cite{yin2025causvid} have further explored efficient causal structures to improve generation speed and interactivity. These works collectively highlight the importance of decoupling sequential units and meticulously modeling their internal correlations in sequence generation tasks.

To this end, we propose a novel method named the \textbf{Temporally Decoupled Diffusion Model (TDDM)}, which aims to model the internal temporal dependencies within a trajectory more finely, thereby generating plans that are more consistent and forward-looking. Our main contributions can be summarized as follows:
\begin{itemize}
    \item We introduce a novel temporally-decoupled training paradigm. We partition the trajectory into multiple temporal segments and apply different levels of noise independently to each segment during training. This randomized temporal masking mechanism significantly enhances the model's ability to capture complex temporal correlations.
    
    \item We design an asymmetric temporal guidance strategy for inference. By extending Classifier-Free Guidance~\cite{Dhariwal2021cfg} to the temporal dimension, our strategy uses weakly noised far-term priors as a condition to guide the generation of a near-term plan. This effectively frames the inference process as a goal-directed trajectory completion task, enhancing the long-term consistency of the plan.
    
    \item We develop an efficient model architecture that supports temporal decoupling. Our architecture, based on the diffusion transformer~\cite{Peebles2023dit} framework, incorporates a Temporally Decoupled Adaptive Layer Normalization (TD-AdaLN) module. This allows for the independent injection of diffusion timestep information for each temporal segment, providing architectural support for our proposed training and inference procedures.
    
    \item Comprehensive evaluations on the large-scale nuPlan~\cite{caesar2022nuplan} benchmark robustly demonstrate the superiority of our proposed model. The results show that TDDM achieves performance that approaches or exceeds that of state-of-the-art learning-based methods, particularly in the most challenging long-tail scenarios, where it exhibits exceptional robustness and planning consistency.
\end{itemize}

\begin{figure*}[t]
\centering
\includegraphics[width=1.0\textwidth]{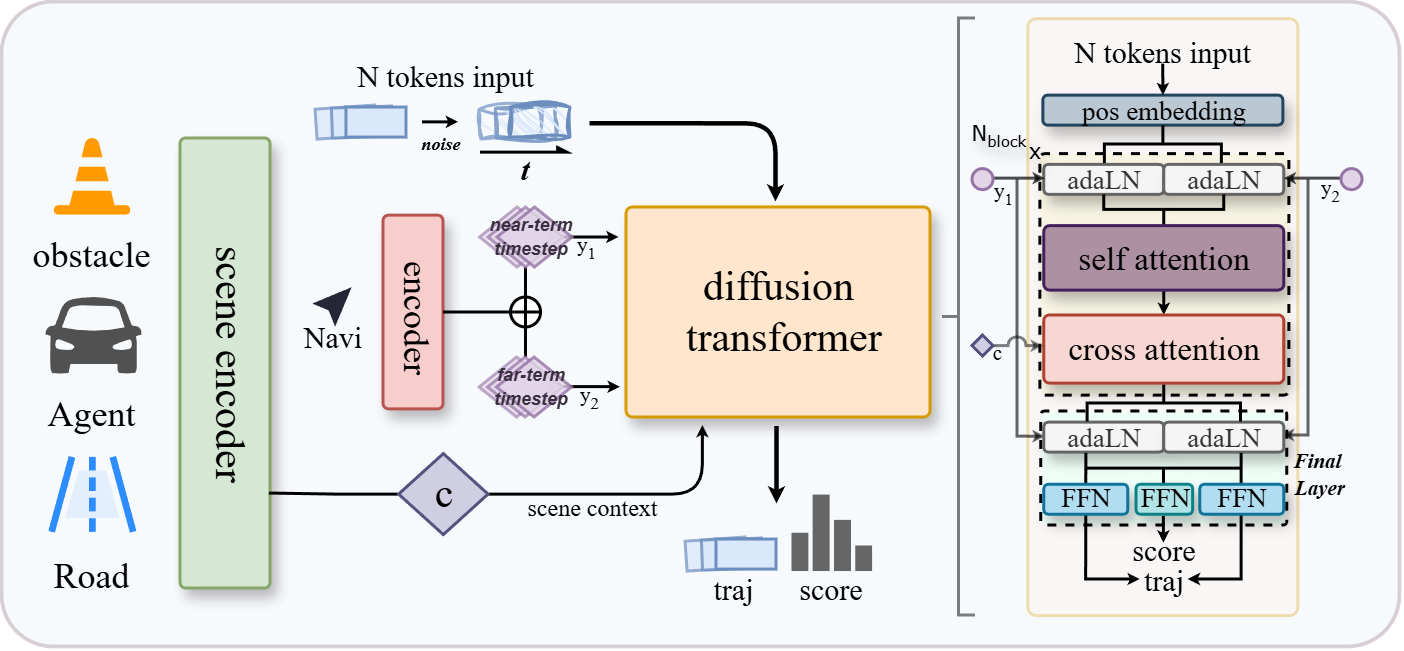} 
\caption{\textbf{Overview of the temporally decoupled diffusion model.} We adopt the diffusion transformer (DiT) architecture as the decoder, with the encoder responsible for encoding environmental context information (including static obstacles, agents, and roads). The decoder can accept independent timesteps from temporal segments through temporally decoupled adaLN.}
\label{fig2}
\end{figure*}

\section{Related Work}

\paragraph{Imitation-based Planning.}
Imitation learning, which learns a policy by mimicking expert demonstrations, is a cornerstone of autonomous driving research. Behavioral Cloning (BC) has been particularly prevalent in end-to-end architectures that map sensor inputs directly to control actions. The architectural evolution of these systems is well-documented: early approaches~\cite{pomerleau1988alvinn} employed simple \textbf{\textit{Convolutional Neural Networks}} (CNN). To capture the temporal nature of driving, subsequent works such as ChauffeurNet~\cite{bansal2018chauffeurnet} incorporated \textbf{\textit{Recurrent Neural Networks}} (RNN). More recently, the \textbf{\textit{Transformer}} architecture has become dominant, with models like planTF~\cite{cheng2023plantf} leveraging its capacity for long-horizon sequence modeling to imitate expert trajectories. To surpass the limitations of vanilla imitation learning, hierarchical approaches like PLUTO~\cite{cheng2024pluto}, which decouples lateral and longitudinal control. A prominent trend is the development of unified end-to-end frameworks~\cite{weng2024drive, jiang2023vad, hu2023planning}, exemplified by UniAD~\cite{hu2023planning}, which integrates perception, prediction, and planning into a single model. However, a persistent challenge for these methods is effectively capturing the multi-modal nature of driving decisions. While some methods, such as VADv2~\cite{chen2024vadv2}, address this by discretizing the action space into a large vocabulary, this challenge has broadly motivated the exploration of more powerful generative paradigms, notably diffusion models~\cite{ho2020denoising, song2021maximum}.

\paragraph{Diffusion Models for Trajectory Planning.}
Diffusion models reframe motion planning as a conditional generation task. They iteratively refine a random noise tensor into one or more feasible trajectories, conditioned on scene context such as sensor data and navigational goals. This paradigm inherently supports the generation of diverse, high-quality outputs and has shown remarkable performance. MotionDiffuser~\cite{jiang2023motiondiffuser} was among the first to apply this concept to multi-agent motion forecasting. Hybrid approaches like Diffusion-ES~\cite{yang2024diffusion} combine diffusion models with evolutionary strategies, using the former to generate a high-quality initial policy population for the latter to optimize, enhancing robustness in uncertain scenarios. To improve efficiency, DiffusionDrive~\cite{liao2025diffusiondrive} initiates the denoising process from learned anchors, accelerating inference and improving planning accuracy. Meanwhile, DiffusionPlanner~\cite{zheng2025diffusionbased} casts planning as a joint prediction problem for all agents and enables style-conditioned trajectory generation. Despite their success, these methods share a fundamental modeling assumption: they treat the entire future trajectory as a monolithic, indivisible entity. This monolithic denoising approach overlooks the heterogeneous nature of temporal dependencies within a trajectory. For instance, near-term waypoints are heavily constrained by immediate collision avoidance, whereas long-term waypoints are dictated by global route adherence. This simplification restricts the model's ability to generate trajectories that optimally balance short-term safety with long-term consistency.

\paragraph{Decoupled Modeling in Sequence Generation.}
Inspiration for addressing this limitation is drawn from recent advances in other sequence generation domains, particularly video synthesis. There, researchers have highlighted the benefits of decoupling sequential units. For example, Diffusion Forcing~\cite{chen2025diffusion} applies noise independently to video frames, forcing the model to learn fine-grained causal relationships. This principle is extended in works like CausVid~\cite{yin2025causvid} and DFoT~\cite{song2025historyguidedvideodiffusion}, which further investigate causal architectures and guided generation on independently noised units. These studies collectively demonstrate that by decomposing a sequence and explicitly modeling inter-unit dependencies, it is possible to significantly improve generation quality and temporal coherence. Motivated by this principle, we introduce the concept of \textbf{\textit{temporal decoupling}} to trajectory planning. We posit that a trajectory should be modeled not as a monolithic whole, but as a sequence of interconnected temporal segments. By subjecting these segments to independent yet coordinated diffusion processes, our model learns a more expressive representation of temporal dependencies. This allows for a more nuanced control over the planning process, enabling our model to reconcile short-term reactivity with long-term goal consistency—a critical capability largely unaddressed by prior monolithic diffusion-based planners.

\begin{figure*}[t]
\centering
\includegraphics[width=1.0\textwidth]{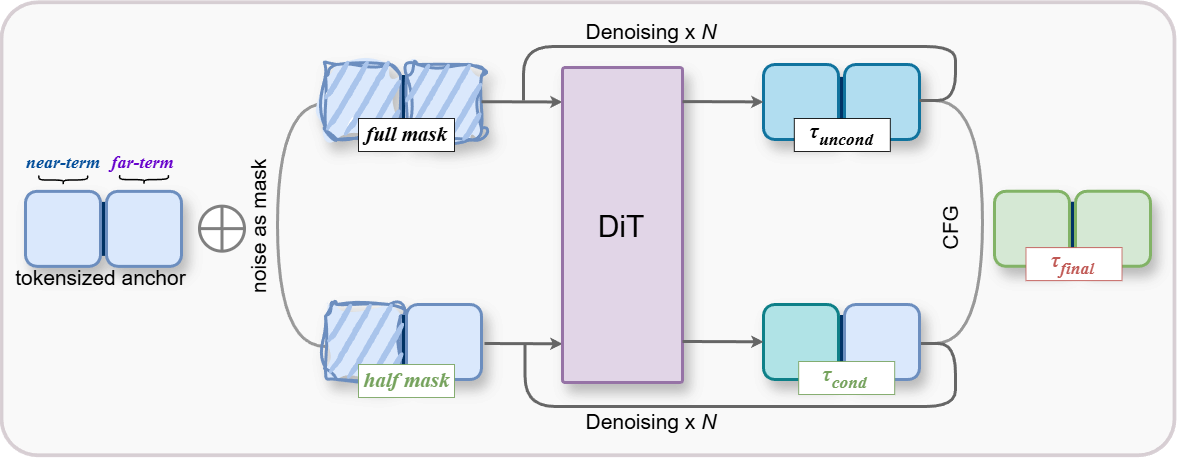} 
\caption{\textbf{Pipeline of the Asymmetric Temporal CFG}. The trajectory is a fusion of outputs from an unconditional path and a conditional path. The unconditional path performs standard full-sequence diffusion. The conditional path, enabled by an asymmetric temporal mask, leverages a nearly clean future prior to guide the denoising generation of the past segment.}
\label{fig3}
\end{figure*}

\section{Preliminary}

\subsection{Conditional Diffusion Model}
A conditional diffusion model learns to reverse a gradual noising process to generate data samples conditioned on some context $c$. The forward process is a fixed Markov chain that incrementally adds Gaussian noise to the original data $\tau^0$ over a series of timesteps. The data distribution at any timestep $i$ can be expressed in a closed form:
\begin{equation}
q(\tau^{i} \mid \tau^{0}) = \mathcal{N}\left( \tau^{i}; \sqrt{\bar{\alpha}^{i}} \, \tau^{0}, (1 - \bar{\alpha}^{i}) \mathbf{I} \right),\label{eq:pre_1}
\end{equation}
where $\tau^i$ is the noised data at timestep $i \in [0, 1]$, and $\{\bar{\alpha}^i\}$ is a predefined noise schedule.

The reverse process involves training a neural network $f_\theta(\tau^i, c, i)$, typically parameterized to predict the added noise, to denoise $\tau^i$ back towards $\tau^0$ given the condition $c$. During inference, starting from pure Gaussian noise $\tau^T$, the model iteratively applies the learned denoising function to generate a clean data sample $\tau^0$. This iterative process can be represented as:
\begin{equation}
p_{\theta}(\tau^0 \mid c) = \int p(\tau^T) \prod_{t=1}^{T} p_{\theta}(\tau^{t-1} \mid \tau^t, c) d\tau^{1:T}.\label{eq:pre_2}
\end{equation}
For practical inference, this continuous-time formulation is approximated by discretizing the time interval into a finite number of steps $T$, where the integral is resolved through an iterative sampling procedure.

\subsection{Anchor-based Trajectory Vocabulary}
To structure the motion planning problem for a generative model, we discretize the continuous action space into a predefined trajectory vocabulary~\cite{liao2025diffusiondrive}, $V = \{v_i\}_{i=1}^{M}$. This approach transforms the problem from regressing a continuous path to selecting and refining the best trajectory prototype from a diverse set.

This vocabulary is created by applying the k-means clustering algorithm to a large-scale dataset of expert-driven trajectories, such as nuPlan~\cite{caesar2022nuplan}. Each resulting cluster centroid $v_i \in V$ serves as a trajectory anchor, representing a distinct, kinematically feasible driving maneuver. Each anchor is a prototype trajectory $\tau$ composed of a sequence of $T_h$ waypoints, where $\tau = \{(x_t, y_t, \phi_t)\}_{t=1}^{T_h}$, capturing the vehicle's position and heading over the planning horizon.

\subsection{Problem Formulation}
Building on the concepts above, we formulate motion planning as a conditional generation task. The core idea is to learn a denoising model that can refine a set of noisy trajectory anchors into a final multimodal trajectory distribution, conditioned on the scene context $c$.

Specifically, we first tokenize each trajectory anchor $v \in V$ by segmenting it into $N$ temporal segments: $\{\tau_1^0, \tau_2^0, \dots, \tau_N^0\}$. During training, we apply independent noise to each segment. For the $n$-th segment $\tau_n^0$, we sample an independent diffusion timestep $i_n \in [0, 1]$ and a standard Gaussian noise $\epsilon_n \sim \mathcal{N}(0, \mathbf{I})$. The resulting noised segment $\tau_n^{i_n}$ is calculated as:
\begin{equation}
\tau_n^{i_n} = \sqrt{\bar{\alpha}^{i_n}} \tau_n^0 + \sqrt{1 - \bar{\alpha}^{i_n}} \epsilon_n.\label{eq:pre_3}
\end{equation}
This independent noising forces the model to learn complex temporal dependencies between segments, rather than relying on uniform noise patterns.

The overall objective is to learn the weights $\theta$ of a denoising network $f_\theta$ that takes the scene context $c$ and the set of noised anchors as input, and outputs a refined set of trajectories $\{\hat{\tau}_k\}$ along with their corresponding confidence scores $\{\hat{s}_k\}$:
\begin{equation}
\{\hat{s}_k, \hat{\tau}_k\}_{k=1}^{M} = f_\theta(\{\{\tau_{k,n}^{i_n}\}_{n=1}^{N}\}_{k=1}^{M}, c).\label{eq:pre_4}
\end{equation}
Essentially, the model learns to jointly denoise all anchors and predict which one represents the optimal plan for the given context.

\section{Approach}
\subsection{Temporally Decoupled Diffusion Model}
We propose the \textbf{Temporally Decoupled Diffusion Model (TDDM)}. Inspired by recent video generation advancements~\cite{chen2025diffusion,song2025historyguidedvideodiffusion} that treat videos as image sequences, we leverage their structural commonalities with trajectory generation. We reformulate the trajectory generation task as a denoising process that is decoupled in the time dimension. Our core idea is to partition a complete trajectory into multiple temporal tokens and apply independent random noise to these tokens during training. This compels the model to not only learn the kinematic smoothness within each token but also to leverage global context to understand and reconstruct the complex temporal correlations between tokens.

To realize our proposed Temporally Decoupled Diffusion Model, the architecture must address two principal challenges: first, it must accommodate the application of distinct diffusion timestep encodings to each temporally decoupled segment; second, it must ensure kinematic consistency across the entire planning horizon. The overall architecture designed for this is shown in Figure \ref{fig2}.

The model's input processing begins with the decomposition of the trajectory into $N$ temporal segments. We conceptually cluster these $N$ segments into $G$ macro-groups. $N$ can be flexibly adjusted, whereas $G$ dictates the instantiation of group-specific architectural components. Figure 2 illustrates our primary experimental setup where $G=2$. Standard group-specific positional encodings are applied to each segment, which are then processed through a shared MLP projection layer, $\mathcal{F}_{\text{pre}}$, mapping each group $\tau_g$ into a partial hidden dimension. These $G$ group features are concatenated along the channel dimension to form a unified feature tensor $h$:
\begin{equation}
h_g = \mathcal{F}_{\text{pre}}(\text{pos}(\tau_g)); \quad h = \text{Concat}(h_1, h_2, \dots, h_G). \label{eq:app_1}
\end{equation}
A key innovation is our \textbf{Temporally Decoupled Adaptive Layer Normalization (TD-AdaLN)} mechanism for injecting the independent diffusion timesteps. For each macro-group $g \in \{1, \dots, G\}$, a conditional vector $y_g$ is constructed by combining its group-specific diffusion timestep encoding $t_g$ with a shared navigation encoding:
\begin{equation}
y_g = \mathcal{F}_{\text{time}}(t_g) + \mathcal{F}_{\text{navi}}(\text{navi}). \label{eq:app_2}
\end{equation}
Here, navigation information $navi \in \mathbb{R}^{(K \times P) \times D_{\text{route}}}$ provides essential route guidance, extracted from the nuPlan benchmark as $K$ route lanes with $P$ points containing $D_{\text{route}}$ coordinate features. Within each transformer block, these conditional vectors are used to generate segment-specific modulation parameters, $\text{params}_g = \mathcal{F}_{\text{adaLN}}(y_g)$. These parameters are then concatenated along the channel dimension to form a complete modulation tensor, $\text{params}$, which is subsequently applied to the main feature tensor $h$ to conditionally modulate its normalization statistics.

The model's backbone consists of a series of DiT blocks where Self-Attention and Cross-Attention modules operate on the complete feature map $h$. Self-Attention facilitates information fusion among different temporal segments, enforcing internal consistency. Cross-Attention allows all segments to integrate unified external scene information $c$. Specifically, $c$ is the fused vectorization of the nearest neighbors (a 2s history of kinematics, size, and category), map elements (polylines, traffic lights, speed limits), and static obstacles. These encoded context features serve as Key/Value pairs in the Cross-Attention layers. Finally, a dedicated feed-forward network (FFN) predicts the confidence score for the whole trajectory, while the segments are decoded independently by their corresponding group-specific FFNs.

\subsection{Asymmetric Temporal CFG}
Temporal decoupling enables unprecedented flexibility in reference. Rather than starting from a single, homogeneous noise state, we can assign distinct denoising starting points to different temporal segments. Based on this, we have designed a novel inference pipeline, as shown in Figure \ref{fig3}.

Our inference process performs two parallel model forward passes at each denoising step. However, unlike traditional CFG, these two passes do not simply represent the "with/without global condition." Instead, they represent two different temporal generation hypotheses:
\begin{enumerate}
    \item \textbf{Unconditional Path}: This path represents the original full-sequence diffusion mode, where we start from a trajectory vocabulary composed entirely of Gaussian noise without temporal decoupling. That is, the same noise is added to all segments, and the noise level is consistent at each denoising step.
    
    \item \textbf{Conditional Path}: Conversely, the Conditional Path defines a deterministic assumption about the long-term future. We divide the trajectory segments into two parts: near-term plan (e.g., the first $N/2$ segments) and far-term plan (e.g., the last $N/2$ segments). At each inference step, we construct an asymmetric, mixed-noise-level input: (1) \textbf{Near-term}: Start from full noise. (2) \textbf{Far-term}: Are always kept as the weakly noised original prototype (anchor) form, corresponding to a diffusion timestep of 0.001. This poses a completion task: given a future target, generate the optimal current path.
    
    \item \textbf{CFG}: After obtaining the outputs from both paths, we fuse them according to the CFG formula:
    \begin{equation}
    \hat{\tau}_{g,\text{final}}^{0} = \hat{\tau}_{g,\text{uncond}}^{0} + w \cdot \left( \hat{\tau}_{g,\text{cond}}^{0} - \hat{\tau}_{g,\text{uncond}}^{0} \right),\label{eq:seg_6}
    \end{equation}
    the guidance scale $w$ controls the adherence strength of the generated result to the long-term target, where $w>1$ enforces stronger conditional guidance and $0<w<1$ weakens it. It is important to note that we actually use the reparameterization trick to directly predict the clean trajectory. 
\end{enumerate}

\begin{table*}[t]
\centering
\small
\renewcommand{\arraystretch}{1.15}
\setlength{\tabcolsep}{8pt}
\begin{tabular*}{\textwidth}{@{\extracolsep{\fill}}llcccccc}
\toprule
\multirow{2}{*}{\textbf{Type}} & \multirow{2}{*}{\textbf{Planner}} 
& \multicolumn{2}{c}{\textbf{Val14}} 
& \multicolumn{2}{c}{\textbf{Test14-hard}}
& \multicolumn{2}{c}{\textbf{Test14}} \\
\cmidrule(lr){3-4} \cmidrule(lr){5-6} \cmidrule(lr){7-8}
 &  & NR & R & NR & R & NR & R \\
\midrule
\textbf{Expert} & Log-replay & 93.53 & 80.32 & 85.96 & 68.80 & 94.03 & 75.86 \\
\midrule
\multirow{6}{*}{\textbf{Rule-based \& Hybrid}} 
& IDM & 75.60 & 77.33 & 56.15 & 62.26 & 70.39 & 74.42 \\
& PDM-Closed & 92.84 & 92.12 & 90.15 & 76.19 & 90.05 & 91.63 \\
& PDM-Hybrid & 92.77 & 92.11 & 65.99 & 70.92 & 90.02 & 91.28 \\
& GameFormer & 79.94 & 79.78 & 68.70 & 70.05 & 79.88 & 82.05 \\
& PLUTO & 92.88 & 86.88 & 80.08 & 76.88 & 92.23 & 90.29 \\
\midrule
\multirow{6}{*}{\textbf{Learning-based}} 
& PDM-Open & 53.53 & 54.24 & 33.51 & 33.89 & 52.81 & 57.23 \\
& UrbanDriver & 68.57 & 64.11 & 50.40 & 49.95 & 51.83 & 67.15 \\
& PlanTF & 84.27 & 76.95 & 69.70 & 61.61 & 85.36 & 79.58 \\
& PLUTO w/o refine. & 88.89 & 78.11 & 70.03 & 59.74 & \underline{89.90} & 78.62 \\
& Diffusion Planner & \textbf{89.76} & \textbf{82.56} & \underline{75.67} & \textbf{68.56} & 89.19 & \textbf{82.55} \\
& TDDM (Ours) & \textbf{89.81} & \underline{79.78} & \textbf{77.95} & \underline{65.20} & \textbf{90.4} & \underline{80.63} \\
\bottomrule
\end{tabular*}
\caption{\textbf{Comparison of planning performance on nuPlan under both non-reactive (NR) and reactive (R) closed-loop evaluation.} Higher scores indicate better performance. Within each method group, the best result is highlighted in \textbf{bold}, and the second best is \underline{underlined}. TDDM is trained on only 200k nuPlan scenarios.}
\label{tab1}
\end{table*}

\begin{table*}[t]
  \centering
  \renewcommand{\arraystretch}{1.15}
  \setlength{\tabcolsep}{6pt}
  \newcolumntype{Y}{>{\centering\arraybackslash}X}
  \label{tab:ablation_components}

  \begin{tabularx}{\textwidth}{c YYYY c}
    \toprule
    \textbf{ID} &
    \textbf{Traj. Token.} &
    \textbf{TD-AdaLN} &
    \textbf{Independent Noise} &
    \textbf{Asymmetric CFG} &
    \textbf{nuPlan (Test14-hard)} \\
    \midrule
    1 & $\times$ & $\times$ & $\times$ & $\times$ & 75.91 \\
    2 & $\checkmark$ & $\times$ & $\times$ & $\times$ & 76.60 \\
    3 & $\checkmark$ & $\checkmark$ & $\times$ & $\times$ & 74.94 \\
    4 & $\checkmark$ & $\times$ & $\checkmark$ & $\times$ & 73.76 \\
    5 & $\checkmark$ & $\checkmark$ & $\checkmark$ & $\times$ & 76.88 \\
    6 & $\checkmark$ & $\checkmark$ & $\checkmark$ & $\checkmark$ & \textbf{77.95} \\
    \bottomrule
    \end{tabularx}
    \caption{\textbf{Ablation study of key components on the nuPlan Test14-hard benchmark.} ID 1 refers to a baseline diffusion planner with an anchor. We analyze the impact of sequentially adding Trajectory Tokenization (Traj. Token.), Temporally Decoupled AdaLN (TD-AdaLN), Independent Noise, and Asymmetric CFG. For Trajectory Tokenization, a consistent number of tokens is used. The model architecture incorporating Temporally Decoupled AdaLN is illustrated in Figure \ref{fig2}.}
    \label{tab:2}
\end{table*}

\begin{figure*}[t]
\centering
\includegraphics[width=0.9\textwidth]{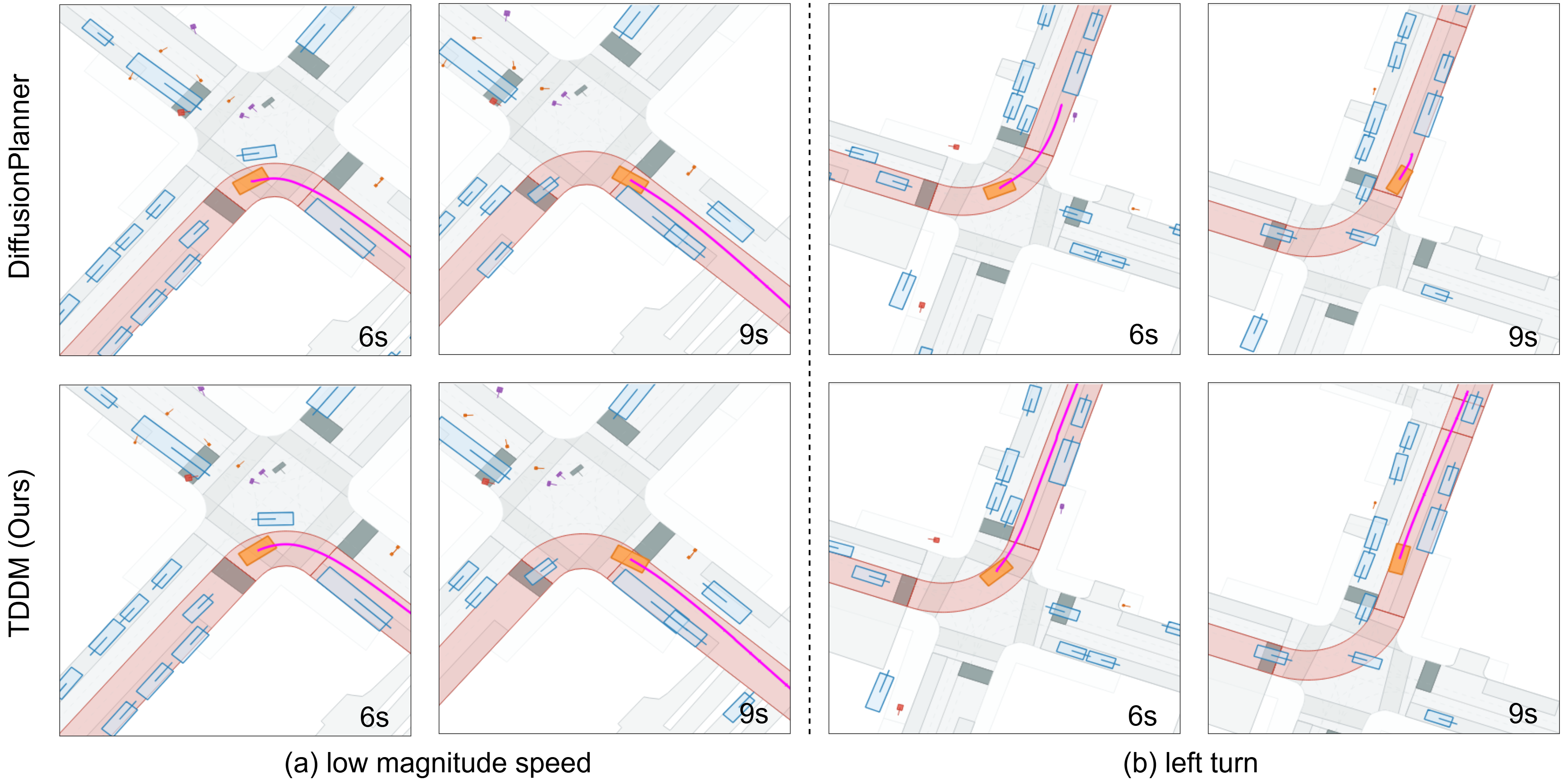} 
\caption{Comparison of Planned Trajectories by Diffusion Planner and our proposed TDDM method across two challenging traffic scenarios. The ego vehicle trajectory is shown in purple.}
\label{fig4}
\end{figure*}

\begin{table}[t]
  \centering
  \renewcommand{\arraystretch}{1.1}
  \setlength{\tabcolsep}{6pt}

  \newcolumntype{Y}{>{\centering\arraybackslash}X}

  \begin{minipage}[t]{0.48\linewidth}
    \centering
    \begin{tabularx}{\linewidth}{YY}
      \toprule
      \makebox[0.48\linewidth][c]{\textbf{Token Num}} &
      \makebox[0.48\linewidth][c]{\textbf{Score}} \\
      \midrule
      1  & 75.55 \\
      2  & 76.41 \\
      4  & \textbf{77.95} \\
      8  & 76.03 \\
      16 & 75.26 \\
      \bottomrule
    \end{tabularx}
    \caption{Token ablation}
    \label{tab:3}
  \end{minipage}
  \hfill
  \begin{minipage}[t]{0.48\linewidth}
    \centering
    \begin{tabularx}{\linewidth}{YY}
      \toprule
      \makebox[0.48\linewidth][c]{\textbf{CFG Scale}} &
      \makebox[0.48\linewidth][c]{\textbf{Score}} \\
      \midrule
      0.75 & 76.04 \\
      1.00 & 77.22 \\
      1.25 & \textbf{77.95} \\
      1.50 & 77.02 \\
      1.75 & 74.70 \\
      \bottomrule
    \end{tabularx}
    \caption{CFG scale ablation}
    \label{tab:4}
  \end{minipage}
\end{table}

\subsection{Training}
Our training process begins by applying independent noise to each temporal segment. These segments are then conceptually divided into groups (e.g., near-term and far-term), where all segments within the same group share a common, randomly sampled diffusion timestep. 

With the necessary components defined in the preliminary section, our training objective is designed as a multi-task learning paradigm. The core idea is not only for the model to reconstruct each part of the expert trajectory but also to assess whether it originates from the optimal trajectory prototype for the current scene, similar to the setup in previous multi-trajectory generation works~\cite{cheng2024pluto,liao2025diffusiondrive}. 
We construct a two-part loss for each prototype: 
(1) high-fidelity reconstruction of the expert trajectory, starting from the trajectory prototype (anchor) that is closest in L2 distance; 
(2) accurate identification of which prototype is the best choice for the current scene.
This joint training strategy is implemented through a hybrid loss function that combines an L1 reconstruction loss and a binary classification loss, simplified as follows:
\begin{equation}
\mathcal{L}(\theta) = \sum\nolimits_{k=1}^{M} [ y_k \mathcal{L}_{\text{rec}}( \hat{\tau}_{k}^{0}, \tau_{gt} ) + \lambda \mathcal{L}_{\text{BCE}}(\hat{s}_k, y_k) ],
\label{eq:seg_5}
\end{equation}
where $\hat{s}_k$ is the predicted confidence score representing the probability of prototype $k$ being the optimal choice for the current scene and $y_k \in \{0, 1\}$ is the corresponding ground-truth label where only one element $y_k$ equals 1.
$\lambda$ is a hyperparameter balancing the two terms.
Since each predicted segment $\hat{\tau}_{k}^{0}$ includes one additional overlapping point to smoothly connect adjacent segments, we further penalize the endpoint mismatch by adding a L1 constraint in the reconstruction term.
\begin{equation}
\begin{split}
\mathcal{L}_{\text{rec}}(\hat{\tau}_{k}^{0}, \tau_{gt}) ={}& ||\hat{\tau}_{k}^{0} - \tau_{gt}||_1 \\
& + \gamma \sum_{g=1}^{G-1} ||\text{end}(\hat{\tau}_{k,g}^{0}) - \text{start}(\hat{\tau}_{k,g+1}^{0})||_1,
\end{split}
\label{eq:app_5}
\end{equation}
where $\text{end}(\cdot)$ and $\text{start}(\cdot)$ denote the last and first waypoints of a group-specific segment, respectively, and $\gamma$ is a hyperparameter to weigh the continuity constraint.

\section{Experiment}

\subsection{Dataset and benchmarks}
We train and evaluate our model on nuPlan~\cite{caesar2022nuplan}, a large-scale multi-modal dataset containing $\sim$1,500 hours of real-world driving data from four cities. nuPlan features a diverse set of urban scenarios, including merging, roundabouts, and complex agent interactions. We test our model in both non-reactive and reactive settings using the following three benchmarks in nuPlan, each covering 14 scenario types: Val14 (1,118 regular validation scenarios), Test14-hard (272 long-tail, high-risk scenarios), and Test14-random (over 200 randomly sampled scenarios). 




We benchmark our method TDDM, against a comprehensive set of baselines to ensure a thorough evaluation. Our comparison includes classic rule-based planners like the IDM~\cite{treiber00phy} and the nuPlan challenge winner PDM~\cite{dauner23apdm} in its various configurations (PDM-Closed, PDM-Open, PDM-Hybrid). Furthermore, we include a range of state-of-the-art learning-based models: the policy-gradient method UrbanDriver~\cite{scheel22urbandriver}; the game-theory-inspired GameFormer~\cite{Huang2023gameformer}; two prominent transformer-based planners, PlanTF~\cite{cheng2023plantf} and PLUTO~\cite{cheng2024pluto}; and DiffusionPlanner~\cite{zheng2025diffusionbased}, another recent diffusion-based approach, which serves as a direct point of comparison for our methodology.

\subsection{Implementation Details}
\paragraph{Architecture and Training.}
For a fair comparison, we align with Diffusion Planner~\cite{zheng2025diffusionbased} by reusing its scene encoder, standard nuPlan perceptual inputs, and most implementation settings. We modify the diffusion backbone into a multi-modal, anchor-based architecture, using 20 trajectory anchors derived via k-means clustering. Due to resource constraints, the model was trained for 500 epochs on a 200k-scenario nuPlan subset using four NVIDIA 3080 GPUs (batch size 320, AdamW optimizer, learning rate $5 \times 10^{-4}$). 

\paragraph{Inference.}
We employ a fast inference strategy with only 2 denoising steps via DPM-Solver++ and a VP noise schedule. In each step, only the highest-confidence trajectory is propagated. Classifier-Free Guidance (CFG) is accelerated through parallel batching. The model outputs an 8 seconds trajectory at 10 Hz.

\subsection{Main Results}
\paragraph{Quantitative result.} Table \ref{tab1} presents a quantitative comparison between TDDM and other state-of-the-art planners on the nuPlan benchmark. This benchmark features two core evaluation modes: Non-Reactive (NR) and Reactive (R). Specifically, in R mode, other agents dynamically react to the ego vehicle, whereas in NR mode, they merely replay historical behaviors. \textbf{Notably}, TDDM achieves competitive results using only 200k scenarios, highlighting its exceptional data efficiency compared to state-of-the-art learning-based methods like Diffusion Planner, which requires 1M scenarios.

In the non-reactive (NR) closed-loop evaluation, the performance of TDDM is on par with, and in some metrics even surpasses, the current state-of-the-art learning-based planners. Specifically, on the general Val14 validation set, TDDM achieves a score of 89.81, which is on par with one of the leading learning-based methods, Diffusion Planner (89.76). The superiority of our methodology becomes particularly evident in more challenging scenarios. On the Test14-hard benchmark, which is specifically designed to evaluate robustness in long-tail scenarios, TDDM obtains a score of 77.95, significantly outperforming Diffusion Planner's 75.67. This result robustly demonstrates the effectiveness of the temporal decoupling mechanism in generating more robust and consistent plans when faced with complex and infrequent events. Furthermore, in the Test14-random test set, TDDM maintains a leading position with a score of 90.4.

Althrough TDDM's score in the reactive (R) setting is slightly below the top method, its strong performance in the non-reactive evaluation is more indicative of the planner's intrinsic quality. This, particularly its excellent results on the difficult benchmark, provides compelling evidence of its advanced planning capabilities.

\paragraph{Qualitative comparison.} 
Figure~\ref{fig4} highlights TDDM's superior planning capabilities in two challenging scenarios where the baseline model fails. In the narrow right turn (a), the baseline Diffusion Planner generates a trajectory that collides with a parked bus. In the left turn with an obstacle (b), it makes a critical misjudgment, stopping unnecessarily behind a parked vehicle. In stark contrast, TDDM successfully navigates both situations by producing smooth, safe, and decisive maneuvers. These results demonstrate that our asymmetric temporal guidance effectively prevents the myopic, inconsistent decisions of the baseline, enabling TDDM to generate trajectories that are both safer and more coherent.

\subsection{Ablation Study}
We ablate key components of TDDM on the nuPlan Test14-hard benchmark (Table~\ref{tab:2}). Our baseline (ID 1), an anchor-based Diffusion Planner, scores 75.91. Simply introducing trajectory tokenization (ID 2), which segments the trajectory without any architectural or training modifications, provides a modest improvement to 76.60, suggesting that representing the trajectory as a sequence of tokens is inherently beneficial. Moreover, we observe an interesting phenomenon that introducing the TD-AdaLN module without independent noise (ID 3) or applying independent noise without the supportive TD-AdaLN architecture (ID 4) degrades performance to 74.94 and 73.76, respectively, while both components are combined (ID 5), the score improves to 76.88. It demonstrates that the TD‑AdaLN module is essential for appropriately handling the segment-specific timestep information produced by the independent noise training. This reflects a critical synergy between the model architecture and the training paradigm, indicating that their alignment is necessary to jointly enhance the learning of robust temporal correlations. Finally, by incorporating our Asymmetric Temporal CFG at inference time (ID 6), the performance sees another significant leap to 77.95, validating the effectiveness of using a coherent far-term plan to guide the generation of the near-term trajectory.

\paragraph{Trajectory Tokenization.}
We investigated the optimal temporal granularity by varying the number of trajectory tokens ($N$). As detailed in Table~\ref{tab:3}, the model's performance exhibits a clear trend, peaking at \textbf{$N=4$} tokens (a 2-second segment length) with a score of \textbf{77.95}. Performance declines when the partition is either too coarse ($N<4$) or too fine ($N>4$). This indicates that a moderate granularity strikes an optimal balance between capturing complex temporal dynamics and maintaining long-term kinematic consistency.

\paragraph{Classifier-free Guidance.}
We ablated the guidance scale $w$ of our Asymmetric Temporal CFG. As shown in Table~\ref{tab:4}, performance peaks at a guidance scale of \textbf{$w=1.25$}. The results reveal a clear trend: scores decrease with either insufficient guidance ($w < 1.25$) or excessive guidance ($w > 1.25$). This indicates that the optimal scale strikes a balance between enforcing long-term goal consistency and preserving short-term reactive flexibility.

\section{Conclusion}
In this work, we introduced the Temporally Decoupled Diffusion Model (TDDM), a novel framework for autonomous driving motion planning that addresses the limitations inherent in monolithic trajectory generation. By reformulating the planning problem through a temporal decoupling paradigm, our model learns to capture the heterogeneous dependencies across a planning horizon. The core contributions—a temporally-decoupled training scheme with independent noise, a supporting TD-AdaLN architecture, and an Asymmetric Temporal Classifier-Free Guidance strategy for inference—work in synergy to produce trajectories that are both reactive to immediate conditions and consistent with long-term goals.

Experiments on the nuPlan benchmark demonstrate that TDDM \textbf{approaches or exceeds} state-of-the-art learning-based methods, showcasing exceptional robustness and coherence on the challenging Test14-hard subset. These results confirm that explicitly modeling distinct temporal structures is vital for high-performance planning in complex urban environments.
\paragraph{Limitiation and future work.}
Despite its strong performance, TDDM's efficacy in fully reactive, closed-loop simulations is an area for improvement. Future work will pursue several promising directions: enhancing agent interaction, potentially by extending our CFG with game-theoretic principles; removing the reliance on a predefined anchor set by exploring autoregressive generation from historical trajectories; and further exploiting the temporal token structure for more explicit and fine-grained trajectory guidance.

\section*{Acknowledgements}
This work is part of the project ``AI Algorithm and Model Development for Future-Oriented Intelligent Driving Systems''
(Grant No. SYG2024087).

\bibliography{aaai2026}

@article{chen2024e2e,
  author={Chen, Li and Wu, Penghao and Chitta, Kashyap and Jaeger, Bernhard and Geiger, Andreas and Li, Hongyang},
  journal={IEEE Transactions on Pattern Analysis and Machine Intelligence}, 
  title={End-to-End Autonomous Driving: Challenges and Frontiers}, 
  year={2024},
  volume={46},
  number={12},
  pages={10164-10183},
  doi={10.1109/TPAMI.2024.3435937}}

@article{chen2025diffusion,
  title={Diffusion forcing: Next-token prediction meets full-sequence diffusion},
  author={Chen, Boyuan and Mart{\'\i} Mons{\'o}, Diego and Du, Yilun and Simchowitz, Max and Tedrake, Russ and Sitzmann, Vincent},
  journal={Advances in Neural Information Processing Systems},
  volume={37},
  pages={24081--24125},
  year={2025}
}

@article{aradi22reinforce,
  author={Aradi, Szilárd},
  journal={IEEE Transactions on Intelligent Transportation Systems}, 
  title={Survey of Deep Reinforcement Learning for Motion Planning of Autonomous Vehicles}, 
  year={2022},
  volume={23},
  number={2},
  pages={740-759},
  doi={10.1109/TITS.2020.3024655}}

@inproceedings{yin2025causvid,
    title={From Slow Bidirectional to Fast Autoregressive Video Diffusion Models},
    author={Yin, Tianwei and Zhang, Qiang and Zhang, Richard and Freeman, William T and Durand, Fredo and Shechtman, Eli and Huang, Xun},
    booktitle={CVPR},
    year={2025}
}

@article{treiber00phy,
  title = {Congested traffic states in empirical observations and microscopic simulations},
  author = {Treiber, Martin and Hennecke, Ansgar and Helbing, Dirk},
  journal = {Phys. Rev. E},
  volume = {62},
  issue = {2},
  pages = {1805--1824},
  numpages = {0},
  year = {2000},
  month = {Aug},
  publisher = {American Physical Society},
  doi = {10.1103/PhysRevE.62.1805},
  url = {https://link.aps.org/doi/10.1103/PhysRevE.62.1805}
}

@InProceedings{dauner23apdm,
  title = 	 {Parting with Misconceptions about Learning-based Vehicle Motion Planning},
  author =       {Dauner, Daniel and Hallgarten, Marcel and Geiger, Andreas and Chitta, Kashyap},
  booktitle = 	 {Proceedings of The 7th Conference on Robot Learning},
  pages = 	 {1268--1281},
  year = 	 {2023},
  editor = 	 {Tan, Jie and Toussaint, Marc and Darvish, Kourosh},
  volume = 	 {229},
  series = 	 {Proceedings of Machine Learning Research},
  month = 	 {06--09 Nov},
  publisher =    {PMLR},
  url = 	 {https://proceedings.mlr.press/v229/dauner23a.html},
}

@inproceedings{cui2019mtp,
author = {Cui, Henggang and Radosavljevic, Vladan and Chou, Fang-Chieh and Lin, Tsung-Han and Nguyen, Thi and Huang, Tzu-Kuo and Schneider, Jeff and Djuric, Nemanja},
title = {Multimodal Trajectory Predictions for Autonomous Driving using Deep Convolutional Networks},
year = {2019},
publisher = {IEEE Press},
url = {https://doi.org/10.1109/ICRA.2019.8793868},
doi = {10.1109/ICRA.2019.8793868},
booktitle = {2019 International Conference on Robotics and Automation (ICRA)},
pages = {2090–2096},
numpages = {7},
location = {Montreal, QC, Canada}
}

@article{pomerleau1988alvinn,
  title={Alvinn: An autonomous land vehicle in a neural network},
  author={Pomerleau, Dean A},
  journal={Advances in neural information processing systems},
  volume={1},
  year={1988}
}

@misc{bansal2018chauffeurnet,
      title={ChauffeurNet: Learning to Drive by Imitating the Best and Synthesizing the Worst}, 
      author={Mayank Bansal and Alex Krizhevsky and Abhijit Ogale},
      year={2018},
      eprint={1812.03079},
      archivePrefix={arXiv},
      primaryClass={cs.RO},
      url={https://arxiv.org/abs/1812.03079}, 
}

@inproceedings{cheng2023plantf,
  author={Cheng, Jie and Chen, Yingbing and Mei, Xiaodong and Yang, Bowen and Li, Bo and Liu, Ming},
  booktitle={2024 IEEE International Conference on Robotics and Automation (ICRA)}, 
  title={Rethinking Imitation-based Planners for Autonomous Driving}, 
  year={2024},
  volume={},
  number={},
  pages={14123-14130},
  doi={10.1109/ICRA57147.2024.10611364}}

@misc{cheng2024pluto,
      title={PLUTO: Pushing the Limit of Imitation Learning-based Planning for Autonomous Driving}, 
      author={Jie Cheng and Yingbing Chen and Qifeng Chen},
      year={2024},
      eprint={2404.14327},
      archivePrefix={arXiv},
      primaryClass={cs.RO},
      url={https://arxiv.org/abs/2404.14327}, 
}

@inproceedings{weng2024drive,
  title={Para-drive: Parallelized architecture for real-time autonomous driving},
  author={Weng, Xinshuo and Ivanovic, Boris and Wang, Yan and Wang, Yue and Pavone, Marco},
  booktitle={Proceedings of the IEEE/CVF Conference on Computer Vision and Pattern Recognition},
  pages={15449--15458},
  year={2024}
}

@inproceedings{hu2023planning,
  title={Planning-oriented autonomous driving},
  author={Hu, Yihan and Yang, Jiazhi and Chen, Li and Li, Keyu and Sima, Chonghao and Zhu, Xizhou and Chai, Siqi and Du, Senyao and Lin, Tianwei and Wang, Wenhai and others},
  booktitle={Proceedings of the IEEE/CVF conference on computer vision and pattern recognition},
  pages={17853--17862},
  year={2023}
}

@inproceedings{jiang2023vad,
  title={Vad: Vectorized scene representation for efficient autonomous driving},
  author={Jiang, Bo and Chen, Shaoyu and Xu, Qing and Liao, Bencheng and Chen, Jiajie and Zhou, Helong and Zhang, Qian and Liu, Wenyu and Huang, Chang and Wang, Xinggang},
  booktitle={Proceedings of the IEEE/CVF International Conference on Computer Vision},
  pages={8340--8350},
  year={2023}
}

@misc{chen2024vadv2,
      title={VADv2: End-to-End Vectorized Autonomous Driving via Probabilistic Planning}, 
      author={Shaoyu Chen and Bo Jiang and Hao Gao and Bencheng Liao and Qing Xu and Qian Zhang and Chang Huang and Wenyu Liu and Xinggang Wang},
      year={2024},
      eprint={2402.13243},
      archivePrefix={arXiv},
      primaryClass={cs.CV},
      url={https://arxiv.org/abs/2402.13243}, 
}

@article{ho2020denoising,
  title={Denoising diffusion probabilistic models},
  author={Ho, Jonathan and Jain, Ajay and Abbeel, Pieter},
  journal={Advances in neural information processing systems},
  volume={33},
  pages={6840--6851},
  year={2020}
}

@article{song2021maximum,
  title={Maximum likelihood training of score-based diffusion models},
  author={Song, Yang and Durkan, Conor and Murray, Iain and Ermon, Stefano},
  journal={Advances in neural information processing systems},
  volume={34},
  pages={1415--1428},
  year={2021}
}

@article{chen2024end,
  title={End-to-end autonomous driving: Challenges and frontiers},
  author={Chen, Li and Wu, Penghao and Chitta, Kashyap and Jaeger, Bernhard and Geiger, Andreas and Li, Hongyang},
  journal={IEEE Transactions on Pattern Analysis and Machine Intelligence},
  year={2024},
  publisher={IEEE}
}

@article{stilman2008planning,
  title={Planning among movable obstacles with artificial constraints},
  author={Stilman, Mike and Kuffner, James},
  journal={The International Journal of Robotics Research},
  volume={27},
  number={11-12},
  pages={1295--1307},
  year={2008},
  publisher={SAGE Publications Sage UK: London, England}
}

@misc{fan2018baidu,
      title={Baidu Apollo EM Motion Planner}, 
      author={Haoyang Fan and Fan Zhu and Changchun Liu and Liangliang Zhang and Li Zhuang and Dong Li and Weicheng Zhu and Jiangtao Hu and Hongye Li and Qi Kong},
      year={2018},
      eprint={1807.08048},
      archivePrefix={arXiv},
      primaryClass={cs.RO},
      url={https://arxiv.org/abs/1807.08048}, 
}

@inproceedings{jiang2023motiondiffuser,
  title={Motiondiffuser: Controllable multi-agent motion prediction using diffusion},
  author={Jiang, Chiyu and Cornman, Andre and Park, Cheolho and Sapp, Benjamin and Zhou, Yin and Anguelov, Dragomir and others},
  booktitle={Proceedings of the IEEE/CVF conference on computer vision and pattern recognition},
  pages={9644--9653},
  year={2023}
}

@inproceedings{yang2024diffusion,
  title={Diffusion-ES: Gradient-free planning with diffusion for autonomous and instruction-guided driving},
  author={Yang, Brian and Su, Huangyuan and Gkanatsios, Nikolaos and Ke, Tsung-Wei and Jain, Ayush and Schneider, Jeff and Fragkiadaki, Katerina},
  booktitle={Proceedings of the IEEE/CVF conference on computer vision and pattern recognition},
  pages={15342--15353},
  year={2024}
}

@inproceedings{liao2025diffusiondrive,
  title={Diffusiondrive: Truncated diffusion model for end-to-end autonomous driving},
  author={Liao, Bencheng and Chen, Shaoyu and Yin, Haoran and Jiang, Bo and Wang, Cheng and Yan, Sixu and Zhang, Xinbang and Li, Xiangyu and Zhang, Ying and Zhang, Qian and others},
  booktitle={Proceedings of the Computer Vision and Pattern Recognition Conference},
  pages={12037--12047},
  year={2025}
}

@inproceedings{
zheng2025diffusionbased,
title={Diffusion-Based Planning for Autonomous Driving with Flexible Guidance},
author={Yinan Zheng and Ruiming Liang and Kexin ZHENG and Jinliang Zheng and Liyuan Mao and Jianxiong Li and Weihao Gu and Rui Ai and Shengbo Eben Li and Xianyuan Zhan and Jingjing Liu},
booktitle={The Thirteenth International Conference on Learning Representations},
year={2025},
url={https://openreview.net/forum?id=wM2sfVgMDH}
}

@misc{song2025historyguidedvideodiffusion,
  title={History-Guided Video Diffusion}, 
  author={Kiwhan Song and Boyuan Chen and Max Simchowitz and Yilun Du and Russ Tedrake and Vincent Sitzmann},
  year={2025},
  eprint={2502.06764},
  archivePrefix={arXiv},
  primaryClass={cs.LG},
  url={https://arxiv.org/abs/2502.06764}, 
}

@misc{caesar2022nuplan,
      title={NuPlan: A closed-loop ML-based planning benchmark for autonomous vehicles}, 
      author={Holger Caesar and Juraj Kabzan and Kok Seang Tan and Whye Kit Fong and Eric Wolff and Alex Lang and Luke Fletcher and Oscar Beijbom and Sammy Omari},
      year={2022},
      eprint={2106.11810},
      archivePrefix={arXiv},
      primaryClass={cs.CV},
      url={https://arxiv.org/abs/2106.11810}, 
}

@InProceedings{scheel22urbandriver,
  title = 	 {Urban Driver: Learning to Drive from Real-world Demonstrations Using Policy Gradients},
  author =       {Scheel, Oliver and Bergamini, Luca and Wolczyk, Maciej and Osi\'nski, B\l{a}\.{z}ej and Ondruska, Peter},
  booktitle = 	 {Proceedings of the 5th Conference on Robot Learning},
  pages = 	 {718--728},
  year = 	 {2022},
  editor = 	 {Faust, Aleksandra and Hsu, David and Neumann, Gerhard},
  volume = 	 {164},
  series = 	 {Proceedings of Machine Learning Research},
  month = 	 {08--11 Nov},
  publisher =    {PMLR},
  url = 	 {https://proceedings.mlr.press/v164/scheel22a.html},
}

@InProceedings{Huang2023gameformer,
    author    = {Huang, Zhiyu and Liu, Haochen and Lv, Chen},
    title     = {GameFormer: Game-theoretic Modeling and Learning of Transformer-based Interactive Prediction and Planning for Autonomous Driving},
    booktitle = {Proceedings of the IEEE/CVF International Conference on Computer Vision (ICCV)},
    month     = {October},
    year      = {2023},
    pages     = {3903-3913}
}

@InProceedings{Peebles2023dit,
    author    = {Peebles, William and Xie, Saining},
    title     = {Scalable Diffusion Models with Transformers},
    booktitle = {Proceedings of the IEEE/CVF International Conference on Computer Vision (ICCV)},
    month     = {October},
    year      = {2023},
    pages     = {4195-4205}
}

@inproceedings{Dhariwal2021cfg,
 author = {Dhariwal, Prafulla and Nichol, Alexander},
 booktitle = {Advances in Neural Information Processing Systems},
 editor = {M. Ranzato and A. Beygelzimer and Y. Dauphin and P.S. Liang and J. Wortman Vaughan},
 pages = {8780--8794},
 publisher = {Curran Associates, Inc.},
 title = {Diffusion Models Beat GANs on Image Synthesis},
 url = {https://proceedings.neurips.cc/paper_files/paper/2021/file/49ad23d1ec9fa4bd8d77d02681df5cfa-Paper.pdf},
 volume = {34},
 year = {2021}
}

@incollection{mitduc2009,
    YEAR       = {2009},
    ISBN       = {978-3-642-03990-4},
    BOOKTITLE  = {The DARPA Urban Challenge},
    VOLUME     = {56},
    SERIES     = {Springer Tracts in Advanced Robotics},
    EDITOR     = {Buehler, Martin and Iagnemma, Karl and Singh, Sanjiv},
    DOI        = {10.1007/978-3-642-03991-1_5},
    TITLE      = {A Perception-Driven Autonomous Urban Vehicle},
    URL        = {http://dx.doi.org/10.1007/978-3-642-03991-1_5},
    PUBLISHER  = {Springer Berlin Heidelberg},
    AUTHOR     = {John Leonard and Jonathan How and Seth Teller and Mitch Berger and
                 Stefan Campbell and Gaston Fiore and Luke Fletcher and Emilio Frazzoli
                 and Albert Huang and Sertac Karaman and Olivier Koch and Yoshiaki
                 Kuwata and David Moore and Edwin Olson and Steve Peters and Justin Teo
                 and Robert Truax and Matthew Walter and David Barrett and Alexander
                 Epstein and Keoni Maheloni and Katy Moyer and Troy Jones and Ryan
                 Buckley and Matthew Antone and Robert Galejs and Siddhartha
                 Krishnamurthy and Jonathan Williams},
    PAGES      = {163-230},
}

\clearpage
\appendix

\section{Algorithms}
In this appendix, we provide the detailed pseudo-codes for both the training and inference phases of our proposed Temporally Decoupled Diffusion Model (TDDM). 
\begin{algorithm}[!ht]
\caption{Training and Inference of TDDM}
\label{alg:tddm}
\small

\textbf{Training Phase} \\
\textbf{Input:} Context $c$, GT trajectory $\tau_{gt}$, anchors $V$, segment num $N$, group num $G$, model $f_\theta$
\begin{algorithmic}[1]
\FOR{each training iteration}
    
    \STATE \textbf{Label Assignment:}
    \STATE Find anchor $k^* \in V$ closest to $\tau_{gt}$ in L2 distance.
    \STATE Set target $y_{k^*} \leftarrow 1$, and $y_k \leftarrow 0$ for all $k \neq k^*$.
    
    \STATE \textbf{Tokenization \& Decoupled Noising:}
    \STATE Tokenize all anchors $v \in V$ evenly into $N$ temporal segments.
    \STATE Partition the $N$ segments sequentially into $G$ macro-groups.
    \FOR{each macro-group $g \in \{1, \dots, G\}$}
        \STATE Sample a shared diffusion timestep $i_g \sim \mathcal{U}(0, 1)$.
        \STATE Sample independent noise $\epsilon_n \sim \mathcal{N}(0, \mathbf{I})$ for each segment $n \in g$.
        \STATE Corrupt each segment $n \in g$ to obtain $\tau_n^{i_g}$ (via Eq. 3).
    \ENDFOR
    
    \STATE \textbf{Denoising Forward Pass:}
    \STATE Predict clean trajectories and confidence scores for all anchors:
    \STATE \quad $\{\hat{\tau}^0, \, \hat{s}\} \leftarrow f_\theta(\{ \tau_n^{i_g} \}_{n=1}^N, \, c)$
    
    \STATE \textbf{Loss Computation \& Update:}
    \STATE Compute L1 reconstruction loss $\mathcal{L}_{\text{rec}}$ for $k^*$ (via Eq. 9).
    \STATE Compute BCE classification loss $\mathcal{L}_{\text{BCE}}(\hat{s}, y)$.
    \STATE Update $\theta$ via gradient descent on $\mathcal{L} = \mathcal{L}_{\text{rec}} + \lambda \mathcal{L}_{\text{BCE}}$.
    
\ENDFOR
\end{algorithmic}

\vspace{0.8em}
\hrule
\vspace{0.8em}

\textbf{Inference Phase} \\
\textbf{Input:} Context $c$, anchors $V$, steps $N_{\text{infer}}$, CFG scale $w$, model $f_\theta$
\begin{algorithmic}[1]
    
    \STATE \textbf{Initialization:}
    \STATE Split all anchors $v \in V$ into Near-term ($g_1$) and Far-term ($g_2$).
    
    \STATE \textbf{Unconditional Path (Full Sequence):}
    \STATE $\tau_{\text{unc}}^T \leftarrow$ Add full Gaussian noise ($T \!=\! 1.0$) to $g_1$ and $g_2$.
    \FOR{$t = T$ \textbf{down to} $0$ via $N_{\text{infer}}$ steps}
        \STATE Predict clean states and scores: $\hat{\tau}^0, \hat{s} \leftarrow f_\theta(\tau_{\text{unc}}^t, c)$.
        \STATE Prune candidates to the single trajectory with max score $\hat{s}$.
        \STATE Update $\tau_{\text{unc}}^{t-1}$ via diffusion solver step.
    \ENDFOR
    \STATE Let $\hat{\tau}_{\text{unc}}^0$ be the final predicted clean trajectory.
    
    \STATE \textbf{Conditional Path (Asymmetric Mask):}
    \STATE $\tau_{\text{cond}}^T \leftarrow$ Full noise ($T \!=\! 1.0$) for $g_1$, weak noise ($T \!=\! 0.001$) for $g_2$.
    \FOR{$t = T$ \textbf{down to} $0$ via $N_{\text{infer}}$ steps}
        \STATE Keep $g_2$ strictly fixed at the weak noise level ($T \!=\! 0.001$).
        \STATE Predict clean states and scores: $\hat{\tau}^0, \hat{s} \leftarrow f_\theta(\tau_{\text{cond}}^t, c)$.
        \STATE Prune candidates to the single trajectory with max score $\hat{s}$.
        \STATE Update $\tau_{\text{cond}}^{t-1}$ for $g_1$ via diffusion solver step.
    \ENDFOR
    \STATE Let $\hat{\tau}_{\text{cond}}^0$ be the final predicted clean trajectory.
    
    \STATE \textbf{Asymmetric CFG Fusion:}
    \STATE $\hat{\tau}_{\text{final}}^0 \leftarrow \hat{\tau}_{\text{unc}}^0 + w \cdot (\hat{\tau}_{\text{cond}}^0 - \hat{\tau}_{\text{unc}}^0)$ (via Eq. 7).
    
    \STATE \textbf{Return} final trajectory $\hat{\tau}_{\text{final}}^0$.
\end{algorithmic}
\end{algorithm}
\section{Detailed Experimental Setup}

\paragraph{Datasets.} We utilize a training subset of 200,000 diverse scenarios from the large-scale nuPlan dataset. To construct the model inputs, we extract continuous driving data into an ego-centric spatial-temporal representation. For the ego vehicle, we extract 2 seconds of historical states (at 10 Hz) and 8 seconds of ground-truth future trajectories, alongside an explicitly computed 10-dimensional instantaneous state. For the surrounding environment, we query elements within a 100-meter radius, retaining the 2-second historical and future trajectories of up to 32 closest dynamic agents (with a strict limit of 10 for vulnerable road users) and 5 static obstacles. Missing agent observations are handled via forward padding. For map topology, we apply a Breadth-First Search (BFS) to repair disconnected route roadblocks, and extract polylines for up to 70 surrounding lanes and 25 route-specific lanes. All categorical features (e.g., semantic types, traffic lights) are encoded as one-hot vectors, and heterogeneous inputs are padded to unified dimensions.

\paragraph{Normalization.} To stabilize the diffusion process, all input modalities are transformed into the ego vehicle's local coordinate system at the current timestep. We apply rigorous standardization to the heterogeneous inputs (ego states, agent histories, and map elements) using predefined dataset statistics, strictly excluding zero-padded masks. we apply z-score normalization to the $x$-axis and scale the $y$-axis by the identical standard deviation to prevent geometric distortion.
\end{document}